\documentclass{article}
\DeclareUnicodeCharacter{202F}{~}
\usepackage{xcolor}
\usepackage{subcaption}
\usepackage{amsmath}

\usepackage{amssymb}

\usepackage{float}
\usepackage{svg}
\usepackage{authblk}
\usepackage{soul}

\usepackage{geometry}
\usepackage{hyperref}
\usepackage{url}
\usepackage{natbib}
\usepackage{overpic}
\usepackage{authblk}

\newcounter{commentcounter}
\usepackage{ulem} 
\usepackage{graphicx} 
\usepackage{subcaption}

\title{Foraging with the Eyes: Dynamics in Human Visual Gaze and Deep Predictive Modeling}
\author{Tejaswi V. Panchagnula}
\affil{Indian Institute of Technology Madras}

\date{July 2025}

\begin{document}

\maketitle

\section*{Abstract}
Animals often forage via L\`evy walks—stochastic trajectories with heavy-tailed step lengths optimized for sparse resource environments. We show that human visual gaze follows similar dynamics when scanning images. While traditional models emphasize image-based saliency, the underlying spatio-temporal statistics of eye movements remain underexplored. Understanding these dynamics has broad applications in attention modeling and vision-based interfaces. In this study, we conducted a large-scale human subject experiment involving 40 participants viewing 50 diverse images under unconstrained conditions, recording over 4 million gaze points using a high-speed eye tracker.  Analysis of these data shows that the gaze trajectory of the human eye also follows a L\`evy walk akin to animal foraging. This suggests that the human eye forages for visual information in an optimally efficient manner. 
Further, we trained a convolutional neural network (CNN) to predict fixation heatmaps from image input alone. The model accurately reproduced salient fixation regions across novel images, demonstrating that key components of gaze behavior are learnable from visual structure alone.
Our findings present new evidence that human visual exploration obeys statistical laws analogous to natural foraging and open avenues for modeling gaze through generative and predictive frameworks.

\section{Introduction}
Understanding the spatiotemporal patterns of human eye movements is central to vision science and human–computer interaction. However, most existing models of visual attention treat gaze as a static prediction problem. For example, Judd et al. \cite{judd2009learning} collected eye-tracking data from 15 viewers on 1003 natural images and trained a model to predict fixation density maps using low- and mid-level image features. Crucially, their work also acknowledged that such bottom-up saliency approaches “do not consider top-down image semantics and often do not match actual eye movements.” By collapsing the sequential scanpath into a static heatmap, these models neglect the inherently dynamic nature of gaze.

More recent efforts, such as the work of Xu et al. \cite{xu2014predicting}, attempt to incorporate higher-level semantic cues. Their three-level model—leveraging pixel-, object-, and semantic-level attributes—showed improved prediction accuracy using the OSIE dataset of labeled images. Observers consistently fixated on semantically salient elements such as faces and text, emphasizing the cognitive aspects of visual exploration. Yet, despite these advances, the dominant modeling paradigm remains static: each fixation is treated as an independent spatial sample, rather than part of a temporally ordered sequence. Standard free-viewing protocols (brief exposures of 2–3 seconds) further bias models toward early, saliency-driven fixations and undersample the longer, exploratory phase of gaze behavior.

This lack of temporal modeling stands in contrast to evidence from movement ecology and statistical physics. Brockmann et al.\cite{brockmann2006scaling} demonstrated that human mobility patterns exhibit power-law step length distributions, while Viswanathan et al.\cite{viswanathan2000levy} showed that Lévy walks—random walks with heavy-tailed step lengths—can be optimal for search in sparse environments. These findings suggest that visual exploration, like spatial foraging, may involve both short local sampling and long-range attentional shifts. Indeed, when allowed to freely view images for extended durations, human scanpaths exhibit scale-free statistics indicative of Lévy dynamics.

In this work, we bridge this gap by analyzing full gaze trajectories recorded from 40 subjects viewing 50 images each, over long viewing intervals of 30 seconds. We find that human scanpaths exhibit power-law step length distributions with exponents in the Lévy regime. This suggests that gaze is not simply a function of low-level saliency or semantic relevance, but is governed by a stochastic yet structured exploration process. To further test the learnability of fixation patterns, we train a convolutional neural network (CNN) to predict fixation heatmaps directly from image content. While the model performs well, particularly for high-density fixation zones, it fails to capture the heavy-tailed saccadic jumps observed in human data—underscoring the limits of current saliency learning frameworks and motivating a deeper integration of temporal dynamics in predictive models.

\section{Experimental Procedure}

The experiment that was conducted to gather gaze tracking data to analyze human visual perception was a human subject-based study with 40 subjects taking part. The data were collected after informed consent was obtained from each subject (See Appendix). The data was collected using an Aurora Smart Eye Tracker, which yields gaze tracking data sampled at 120Hz. This gave rise to a time series dataset comprising the coordinates of the location of human eye gaze over the images. 

The experimental setup comprises a regular computer monitor of 1920 by 1080 pixels. Each participant is seated in front of the monitor and an Aurora Smart eye tracker is attached using a magnetic strip to the bottom part of the screen. Subjects were seated comfortably at a consistent viewing distance, with natural head movements allowed for free viewing. 

The data set consists of 50 images. There are a combination of paintings, real world scenes, and abstract art as part of the dataset. The motivation for inclusion of paintings as part of the images is that we find on many occasions that art has the power to direct the human gaze in a certain fashion. This raises a key question: are certain images inherently gaze-directing, or are we neurologically predisposed to view them similarly? One of the objectives of this study is to quantitatively analyze the effect of each of these possibilities.

Images were presented in a slideshow format, where each image is on the screen for 30 seconds. There is a 5 second black blank screen in between each image and at the beginning and the end of the stimulus. The data set shown to the participants is divided into two groups: Group 1 and Group 2. Each group consists of 25 images. The two groups of images were matched using their entropy distribution, since entropy has been shown to be a good measure of the information content in an image (adding reference). 

Based on the timings of each image, the total duration of the stimulus is 15 minutes and 50 seconds long. This was the main reason for splitting the subjects into two groups as viewing the entire 50 images by one person at a time is not possible. Thus, each subject had only 27 images in his/her stimulus.The experiment protocol was refined using participant feedback. Data collected from subjects were stored securely, with strict confidentiality, and details about participants are not released in the public domain. 

On average 110,000 data points were collected per subject over the course of his/her trial. This brings the total number of data points including all participants to around 4 million. A data set such as this is very rare and can help provide many interesting discoveries and insights into the visual perception process of the human brain. 

\section{Data Visualization}

The data recorded in the experiments are quite extensive. Different metrics are present in the data, ranging from velocity profiles, number of stimuli, to the position of the head with respect to the sensor. In this study, the main focus is going to be on the gaze trajectories, specifically the Gaze X coordinates and the Gaze Y coordinates. We find throughout the scientific literature that the coordinates themselves seldom find recognition, but are rather clubbed together as heat maps. The authors believe that the actual gaze trajectory is hiding within itself, an abundance of quantitative insights to be discovered. 

\begin{figure}
    \centering
    \includegraphics[width=1\linewidth]{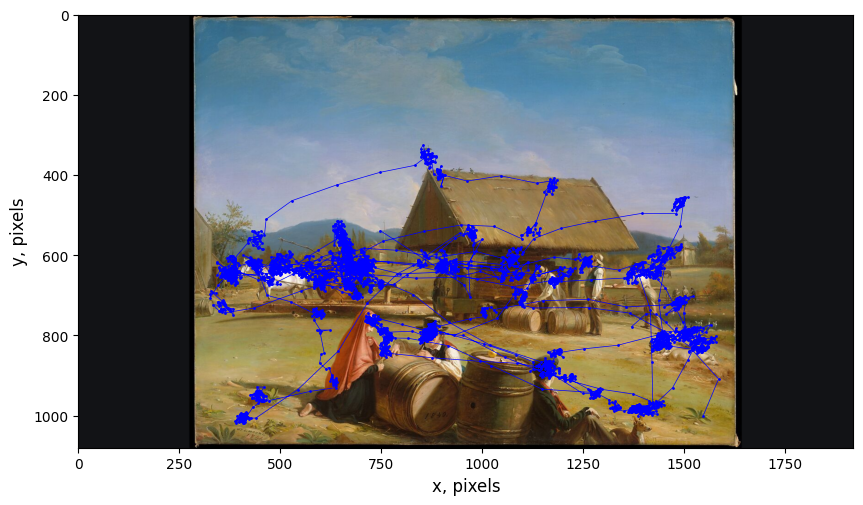}
    \caption{Gaze trajectory of a subject over an image. We can notice the fixation points quite clearly. They cover all the regions of interest in the image which contain relevant information}
    \label{fig:Figure 1}
\end{figure}

A basic visualization of the trajectory over a random image is shown in Figure 1 \ref{fig:Figure 1}. From a cursory perspective, we find that the trajectory follows a very "ruthless" approach. Regions of high information content as surveyed by the brain are the regions where fixations occur. The eye gaze seems to follow a pattern of fixating at one place which contains information and performing fixational eye movements\cite{martinez2013impact}(tremor, microsaccades, and drift) and then moving to the next region of interest via saccades. In general, the gaze of the eyes seems to follow an interesting pattern. In this study, our objective is to quantitatively define a distribution for the eye gaze trajectory and construct a paradigm for generating the trajectory on new images. One must bear in mind that all the trajectories are temporal in nature and that the human eye is fixating on regions in a temporal fashion. This means that we first look at a random position in the image and then move on to the next most "interesting" regions.

We can now visualize the trajectory taken by each subject's eyes over any given image. Preliminary investigations into this area show that not all trajectories are identical in the regions of interest. Of course, the major regions of high information content are seen first by all subjects, but then we see that the trajectories of the eye gaze begin to digress. Possible reasons for this nonconformity in the trajectories are being investigated, some being differences in the background of the subjects: gender, age, etc.

A basic metric of information content in an image is the entropy of the image\cite{attneave1954some}. Shannon's entropy\cite{wu2013local} has been widely used in this regard to determine the entropy of images. As mentioned earlier, we can see a variety of images in this dataset, some with lower levels of entropy and some on the higher side. The entropy ($H$) of the images is calculated as:
\[
H = - \sum_{i=0}^{L-1} p(i) \log_2 p(i)
\]
where L is the number of discrete intensity levels of the pixels and $p$ is their probability distribution.

\begin{figure}[ht]
    \centering
    \begin{subfigure}[t]{0.49\linewidth}
        \centering
        \includegraphics[width=\linewidth]{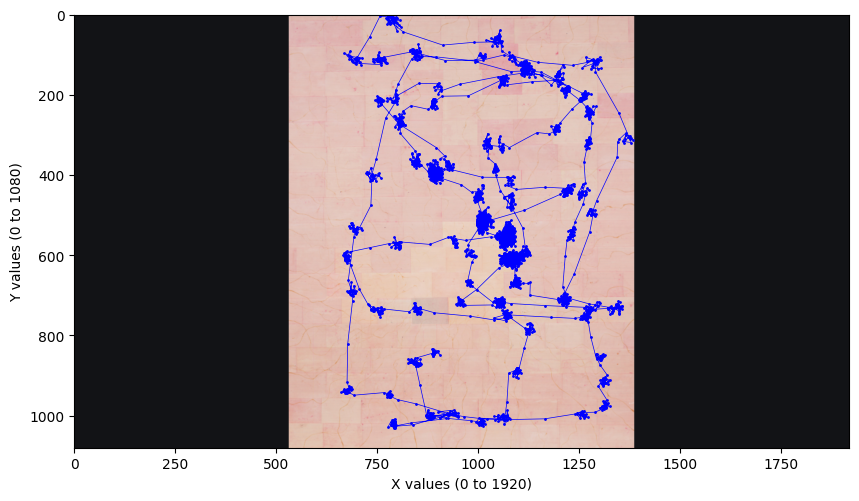}
        \caption{Low Entropy Image}
        \label{fig:low-entropy}
    \end{subfigure}
    \hfill
    \begin{subfigure}[t]{0.49\linewidth}
        \centering
        \includegraphics[width=\linewidth]{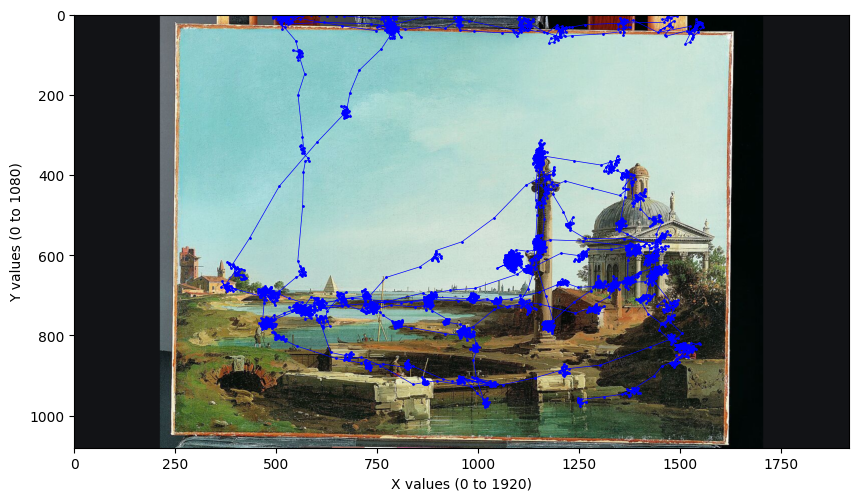}
        \caption{High Entropy Image}
        \label{fig:high-entropy}
    \end{subfigure}
    \caption{Comparison of gaze trajectories over two images: (a) low entropy and (b) high entropy. Despite differing entropy levels, the gaze patterns show no significant qualitative difference across subjects.}
    \label{fig:entropy-comparison}
\end{figure}

In Figure  \ref{fig:entropy-comparison}, we show a comparison between the gaze trajectories of images with low and high entropy. While no stark qualitative differences are immediately visible, subtle distinctions may still exist in the structure and dynamics of the trajectories. To uncover these patterns, we now turn to a quantitative analysis of the gaze data using statistical and clustering techniques.

\section{Statistical Analysis}

Given the abundance of data from different subjects, a preliminary statistical analysis is performed with two metrics. A step length distribution of both the cumulative data (all subjects and all images) is first examined. Subsequently, two conditional distributions, each conditioned on a given individual as well as a given image are calculated. 

\subsection{Step Length Distributions}
 First we compute the step length distribution. Step length is defined as the Euclidean distance between any ($x_i$, $y_i$) and ($x_{i+1}$, $y_{i+1}$). We can now plot the step length distribution for the entire dataset. The step length distribution is an insightful graph in the sense that it yields information on what kind of distribution the eye gaze trajectory follows. 

\subsubsection{Combined Dataset}

\begin{figure}[ht]
    \centering
    \begin{overpic}[width=1\linewidth]{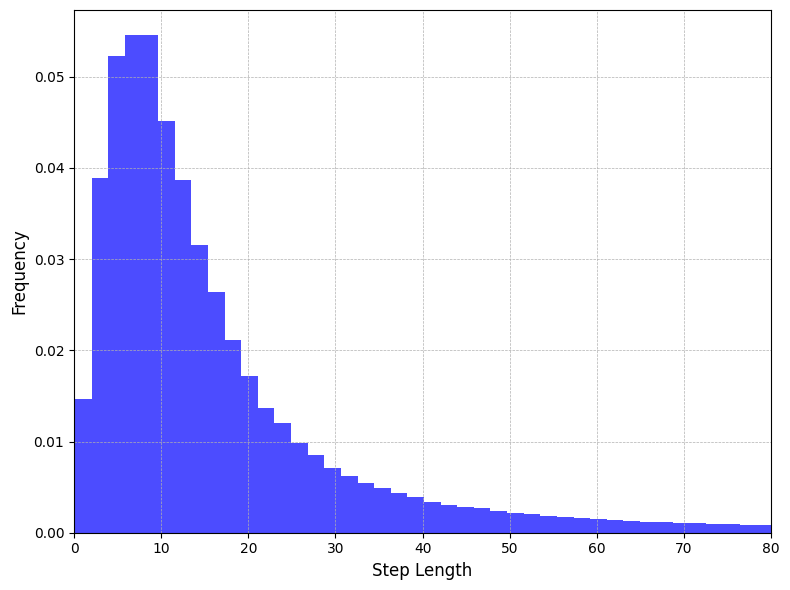}
        \put(37,27){%
            \includegraphics[width=0.55\linewidth]{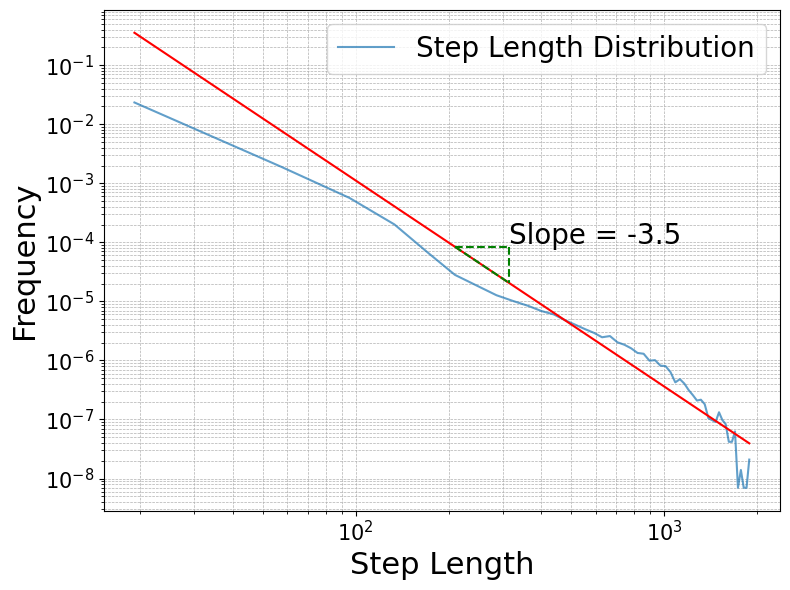}
        }
    \end{overpic}
    \caption{Cumulative Step Length Distribution of all subjects on all images. The mode of the distribution occurs at the 8-10 pixel mark. The inset figure in the top right shows the step length distribution of the combined dataset on a log-log scale, where the linear slope of the tail is approximately -3.49, indicating a heavy-tailed distribution.}
    \label{fig:combined-steplength}
\end{figure}

Figure 3 \ref{fig:combined-steplength} is the frequency distribution of all the step lengths taken by the eyes of every subject in every image. As expected, one can see that the majority of movements by the eyes are within the 100-pixel mark. This is because of the fact that the eyes make small fixational eye movements when observing a scene which make up the bulk of the eye movements by number, given the fact that in each fixation itself there are many of these small step length movements that occur.

Looking at the graph in Figure 3 \ref{fig:combined-steplength}, which is essentially a zoomed view of the same combined frequency plot, it is evident that the maximum number of eye movements' step lengths is in the 8-10 pixel range. This is due to the large number of fixational points in the images shown which are known as microsaccades. From these data, we can conclude that most of the microsaccade lengths are in the 2-20 pixel range. 

The frequency plot of the step lengths, when plotted on a log-log scale (inset in Figure 3\ref{fig:combined-steplength}) clearly shows a linear decay which shows that the step length distribution follows a power law decay, which suggests a fatter tail than a Gaussian distribution. Fitting a trend line for the tail of the distribution shows an average slope of -3.5.  Such a distribution can be characterized as a random walk, which is one of the paths taken during the pursuit of optimal foraging. As the slope of the log-log graph is -3.5, this can be characterized as a Gaussian random walk\cite{viswanathan2000levy}. Usually normal distributions do tend to occur in such circumstances where there are a large number of data points involved. Therefore, one can conclude that this distribution is a heavy tailed distribution. Further statistical analysis is to be done to understand the exact nature of the distribution.

\subsubsection{Image- and Subject-Conditioned data}
\begin{figure}[ht]
    \centering
    \begin{subfigure}[t]{0.49\linewidth}
        \centering
        \includegraphics[width=\linewidth]{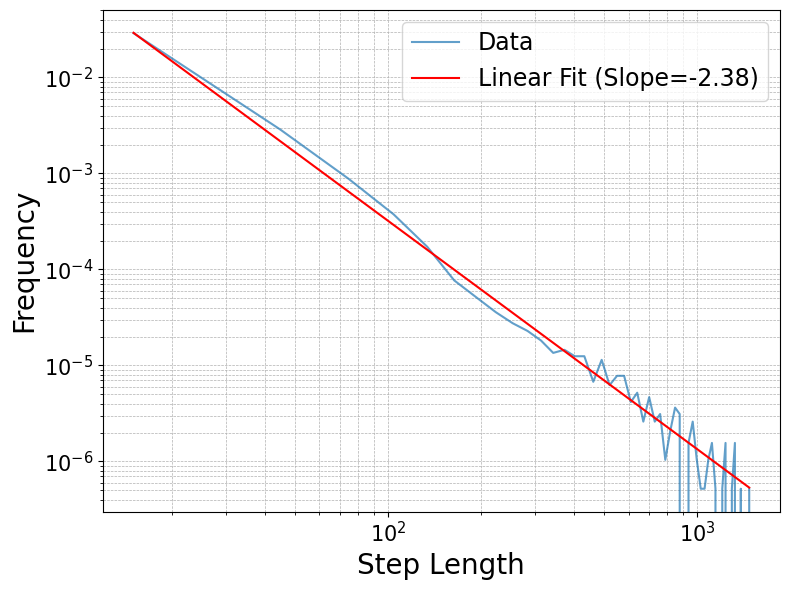}
        \label{fig:all-subjects}
    \end{subfigure}
    \hfill
    \begin{subfigure}[t]{0.49\linewidth}
        \centering
        \includegraphics[width=\linewidth]{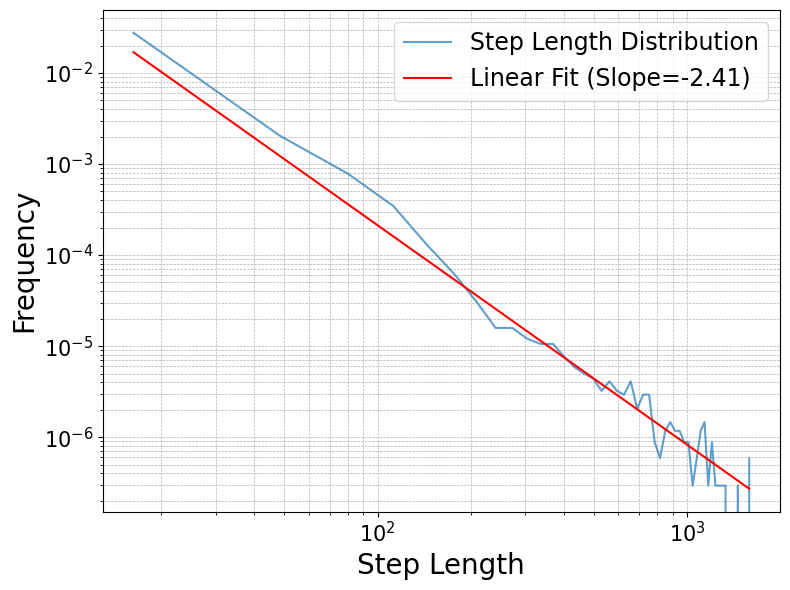}
        \label{fig:one-subject}
    \end{subfigure}
    \caption{Step length distributions: (a) All subjects conditioned over one image, the approximated slope of the graph is -2.38 and (b) one subject conditioned over all images, the approximated slope of the graph is -2.41. Both graphs' approximate slopes lie within the limits for a Levy Walk distribution.}
    \label{fig:side-by-side}
\end{figure}

One of the main findings of this work lies in the distribution pattern conditioned over each image individually. Given the dataset, we know that around 40 trajectories are present for each image. Analyzing the step length distribution for each image over all the subjects, we find that it is very much similar to the overall combined distribution. It is also a heavy tailed distribution but the slope of the linear decrease on the log-log graph is lesser than the combined distribution. Figure \ref{fig:side-by-side} shows that the slope of the log-log graph of the step length distribution of all the subjects over a single image is -2.2. This means that this too is a heavy-tailed distribution, specifically it is a L\`evy walk \cite{viswanathan2000levy}\cite{viswanathan2008levy}.  The interesting result we find is that all the distributions of each image follow a power law decay with $1 < \mu \leq 3$  range, where $\mu$ is the slope in the log-log graph. This indicates that all these distributions are L\`evy walk distributions.

The equation for a Lévy walk in one dimension can be expressed as:

\[
x(t) = \sum_{i=1}^N l_i
\]

where \( l_i \) are steps drawn from a probability distribution with a power-law tail:

\[
P(l) \sim |l|^{-\mu}, \quad 1 < \mu \leq 3.
\]

Here: \( x(t) \) is the position at time \( t \), \( l_i \) are the random step lengths and \( \mu \) is the Lévy distribution parameter. This implies that longer steps occur with lower frequency but are still non-negligible, which is a key property of Lévy processes.

Although we find that the cumulative step length distribution of all the images together is similar to a Gaussian Random Walk, each individual image's distribution is Levy in nature. This does not come as a surprise as the cumulative distribution of all the images put together is essentially a superposition of these individual image trajectories. This gives rise to a Gaussian distribution as a special case of the Central Limit Theorem as all the distributions are independent and identically distributed.

\begin{figure}
    \centering
    \includegraphics[width=0.8\linewidth]{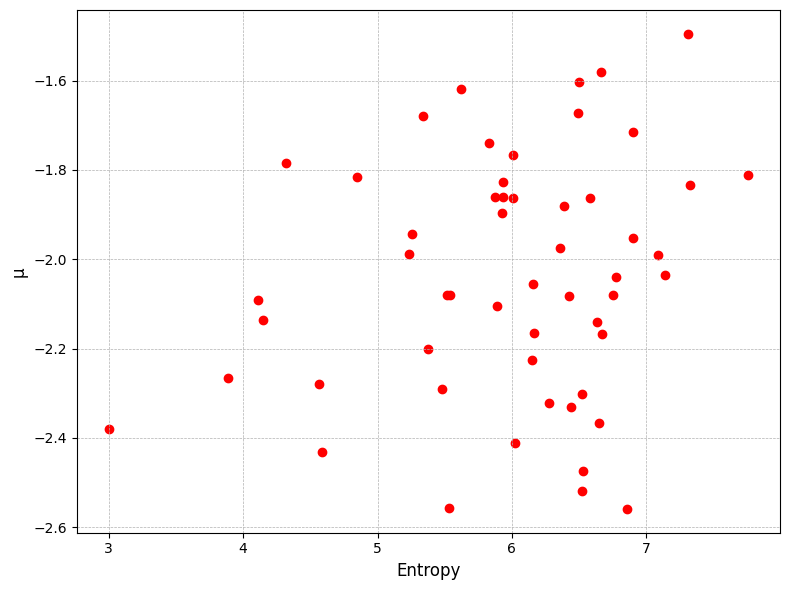}
    \caption{This is a plot of the entropy of the image vs the Levy Walk parameter $\mu$. We can observe that there is a weak positive correlation between the parameters}
    \label{fig:slope_entropy}
\end{figure}

In Figure \ref{fig:slope_entropy}, we plot the entropy of the image versus the approximated slope of the log-log step length distribution. We can observe a slight positive correlation between the two variables, which means images with higher entropy values have a higher chance of larger step lengths. This may be due to the spread of information across the image. A higher entropy indicates larger information content which may mean more regions of fixation. The human eye must then take larger step lengths to encompass all these regions.

\subsection{L\`evy walk is the Optimal Foraging Path}
L\`evy walk is a phenomenon that occurs in nature quite often. Studies have shown that albatross migration\cite{viswanathan1996levy}, dollar bills\cite{brockmann2006scaling} and amoeba movements all follow a Levy walk. The fact that our eyes perform Levy walks to `forage' for visual information is therefore not surprising. Information is the food for our brains and our eyes optimize the foraging for information in the images that they see. L\`evy walks have been shown to be the optimal walk path in foraging. The gaze trajectory of our eyes follows this as, we have many short steps which are the microsaccades that occur during fixation and longer step that occur less frequently which are the saccades from one fixation point to the next. Therefore from the data, we observe that the eyes are performing optimal foraging paths over the image. This in itself is an interesting  result. 

\begin{figure}[ht]
    \centering
    \begin{subfigure}[t]{0.49\linewidth}
        \centering
        \includegraphics[width=\linewidth]{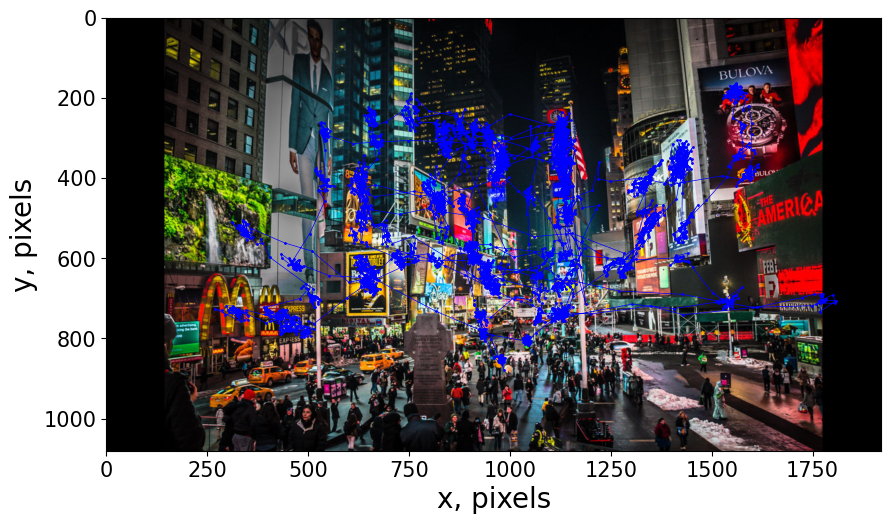}
        \caption{Gaze trajectory over an image with a high $\mu$ coefficient.}
        \label{fig:high-mu}
    \end{subfigure}
    \hfill
    \begin{subfigure}[t]{0.49\linewidth}
        \centering
        \includegraphics[width=\linewidth]{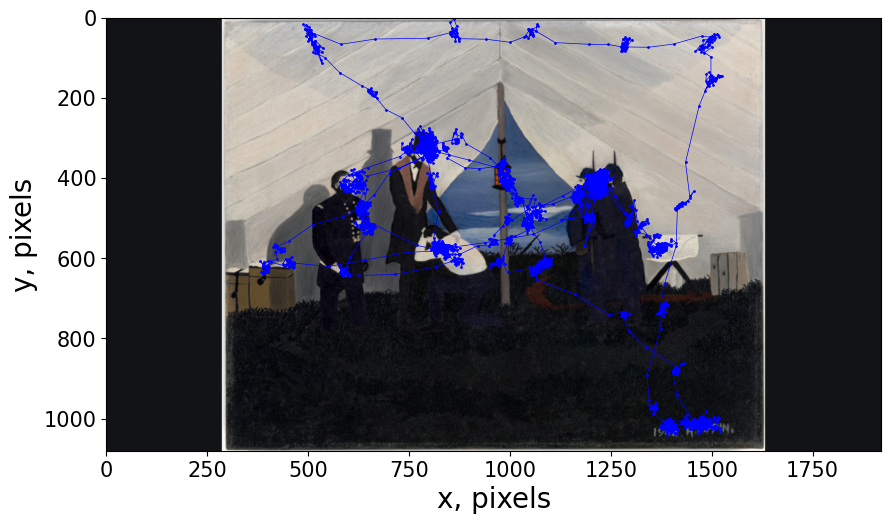}
        \caption{Gaze trajectory over an image with a low $\mu$ coefficient.}
        \label{fig:low-mu}
    \end{subfigure}
    \caption{Comparison of gaze trajectories for images with different $\mu$ coefficients: (a) high $\mu$, and (b) low $\mu$.}
    \label{fig:mu-comparison}
\end{figure}

In Figure \ref{fig:mu-comparison}, which compares the gaze trajectory of a high $\mu$ coefficient versus a lower coefficient. In a L\`evy walk the coefficient essentially controls the step lengths. A lower value of the coefficient will result in larger step lengths, whereas a higher values result in more compact trajectories, in the sense that the steps taken from one fixation to another will be shorter. Qualitatively looking at the images which have higher $\mu$ and those which have a lower $\mu$ an observation that can be made is that images where the regions of interest are more spread out results in a lower L\`evy walk coefficient as the eyes need to take larger steps to explore all the regions of interest. On the other hand, a higher coefficient results from all the regions of interest or fixation points being close together or in one part of the image shown to the subject.

\subsection{Turning Angle}

Our eyes are never still; they constantly move, making rapid and small adjustments that we have captured in the recorded trajectories. A key parameter that offers deeper insight into these movements is the turning angle, which, along with step length, forms a fundamental descriptor of the gaze trajectory's statistical structure. The turning angle helps us understand whether there is any directional preference in eye movements—is every angle equally likely, or do we observe biases?

\begin{figure}[ht]
    \centering
    \includegraphics[width=0.8\linewidth]{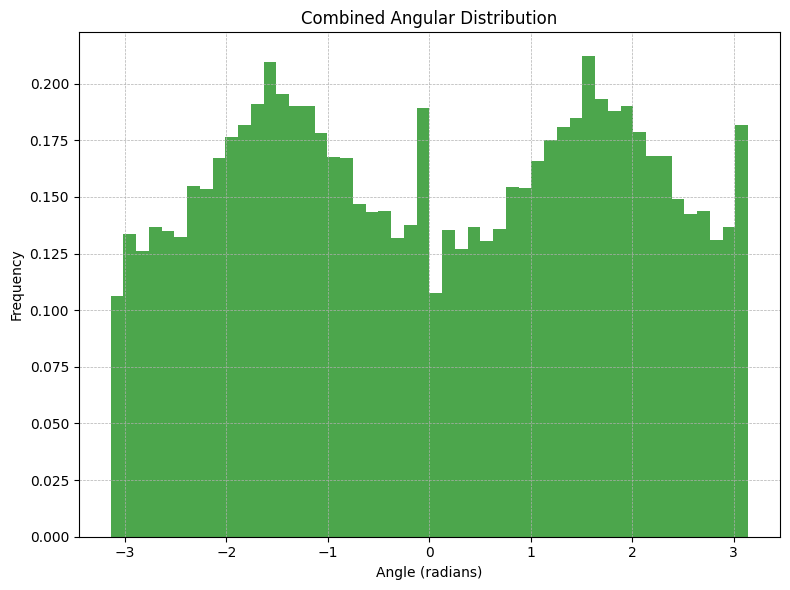}
    \caption{Turning angle distribution of gaze movements across the entire dataset. The x-axis is in radians, ranging from $-\pi$ to $\pi$. Peaks around $\pm \frac{\pi}{2}$ and sharp spikes at $0$ and $\pm \pi$ indicate directional structure in human eye movement.}
    \label{fig:angle-distribution}
\end{figure}

We define the turning angle as the angle subtended at a point $B$ between three successive gaze points $A$, $B$, and $C$; that is, angle $ABC$. The x-axis of the turning angle distribution is expressed in radians, ranging from $-\pi$ to $\pi$. A turning angle of zero indicates that the eye continued in a straight direction, without changing course.

Figure~\ref{fig:angle-distribution} shows the frequency distribution of turning angles computed over the entire dataset. This plot yields a rather counterintuitive observation: the distribution is bimodal, with prominent peaks near $\pm \frac{\pi}{2}$ (i.e., $\pm90^\circ$), indicating a mild preference for vertical shifts. Interestingly, we also observe spikes at angles near $0$ and $\pm\pi$, implying a tendency for straight-line gaze movements and occasional near-complete directional reversals.

These findings suggest two key behaviors: First, the prominent spike at zero radians indicates a preference for straight-line saccadic movements, where the gaze proceeds without significant directional change. Second, the smaller peaks near $\pm\pi$ may correspond to rapid reversals or jitter during fixational periods, where the eye momentarily resets or shuffles within a localized region.

Together, this angular analysis supports a visual exploration model that is both {efficient and structured}, balancing direct saccades with local scanning behaviors.

\section{Modeling and Prediction}

Following our statistical analysis of the gaze data, we proceeded to develop predictive models that estimate regions of visual fixation given a static image as input. The modeling goal was to replicate the spatiotemporal characteristics of human gaze behavior; characterized by sparse and clustered fixations, long-range saccades, and subject-dependent patterns, using deep neural networks.

Initial attempts involved directly predicting the sequence of gaze points over time using recurrent neural networks (RNNs), convolutional LSTMs, and transformer-based architectures. However, these models consistently underperformed. In particular, the predicted sequences tended to drift toward the image center, collapsed into repetitive loops, or converged to local minima lacking the characteristic long-range jumps observed in human gaze. This degradation in quality is likely due to a combination of factors: the intrinsic non-stationary nature of human eye movements, the heavy-tailed nature of step lengths (resembling L\`evy walks), and the high variability in fixation order and saliency prioritization from subject to subject. Because no two human trajectories are alike, and because subjects do not scan images in uniform temporal patterns, the data lack consistent structure in the time domain. Consequently, attempts to model future gaze positions based on past observations using autoregressive techniques failed to produce stable or generalizable results.

To overcome these challenges, we shifted our approach from predicting precise gaze trajectories to estimating where on the image subjects are most likely to look. The model generates a probability map over the visual scene, highlighting potential fixation regions. This strategy reflects the variable and often non-sequential nature of human attention more faithfully.

\subsection{Model Architecture}

To implement this formulation, we developed a lightweight encoder--decoder convolutional neural network based on MobileNetV2 as a backbone. The model follows a U-Net-inspired architecture, with spatial downsampling via MobileNet's convolutional blocks and subsequent upsampling through transposed convolutional layers to recover a dense fixation map. The encoder backbone (MobileNetV2) was initialized with ImageNet-pretrained weights and frozen during early training to prevent overfitting

Let $\mathbf{I} \in \mathbb{R}^{3 \times 224 \times 224}$ denote an input stimulus image. The encoder transforms this input into a deep feature tensor $\mathbf{F} \in \mathbb{R}^{C \times H' \times W'}$, where $C$ is the number of channels. The decoder then upsamples this feature representation to output a single-channel heatmap $\hat{\mathbf{H}} \in \mathbb{R}^{1 \times 112 \times 112}$, with pixel values representing predicted fixation probability densities.

The overall mapping is defined as:
\[
\hat{\mathbf{H}} = D(E(\mathbf{I}))
\]
where $E(\cdot)$ is the MobileNet-based encoder and $D(\cdot)$ is the decoder composed of transposed convolutions.

\subsection{Training the model}

The model was trained using a composite loss function that combines pixelwise reconstruction accuracy with distributional fidelity:

\begin{equation}
\mathcal{L} = \alpha \cdot \text{BCE}(\hat{H}, H) + \beta \cdot \text{MSE}(\hat{H}, H) + \gamma \cdot D_{\text{KL}}(H \parallel \hat{H})
\end{equation}

where:
\begin{itemize}
  \item $\text{BCE}$ is the binary cross-entropy loss,
  \item $\text{MSE}$ is the mean squared error,
  \item $D_{\text{KL}}$ is the Kullback–Leibler divergence between the predicted and ground truth fixation heatmaps,
  \item $H$ denotes the ground truth heatmap constructed from recorded fixation coordinates,
  \item $\hat{H}$ denotes the predicted heatmap.
\end{itemize}

The weights were set to $\alpha = 0.4$, $\beta = 0.3$, and $\gamma = 0.3$ to balance reconstruction accuracy and distributional alignment.

Ground truth heatmaps were generated by convolving gaze coordinates with a 2D Gaussian kernel and downsampled to a resolution of $112 \times 112$. The model training was performed using the AdamW optimizer with cosine annealing over 10 epochs. A subject-balanced train-validation split was used to ensure generalizability across viewers with diverse fixation patterns.

\subsection{Prediction Results and Visualization}

\begin{figure}[ht]
    \centering
    \includegraphics[width=0.95\linewidth]{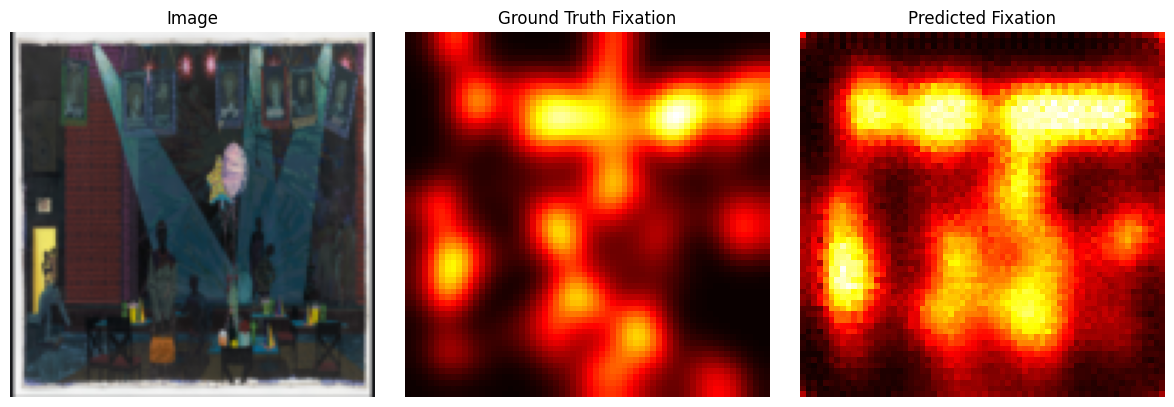}
    \caption{Comparison between predicted and ground truth fixation heatmaps for representative test samples. Predicted maps preserve the spatial structure and intensity gradients of true gaze distributions.}
    \label{fig:heatmap_comp}
\end{figure}

The heatmap-based fixation model was trained for 10 epochs using 85\% of the data for training and 15\% for validation. The model produced consistently high-quality fixation density maps across diverse visual stimuli. Predicted heatmaps exhibited clear multimodal structure, accurately localizing high-attention regions while preserving separation between spatially distinct clusters (Figure~\ref{fig:heatmap_comp}). The predictions showed robustness across image types, capturing both central and peripheral areas of interest.

\begin{figure}[ht]
    \centering
    \includegraphics[width=1\linewidth]{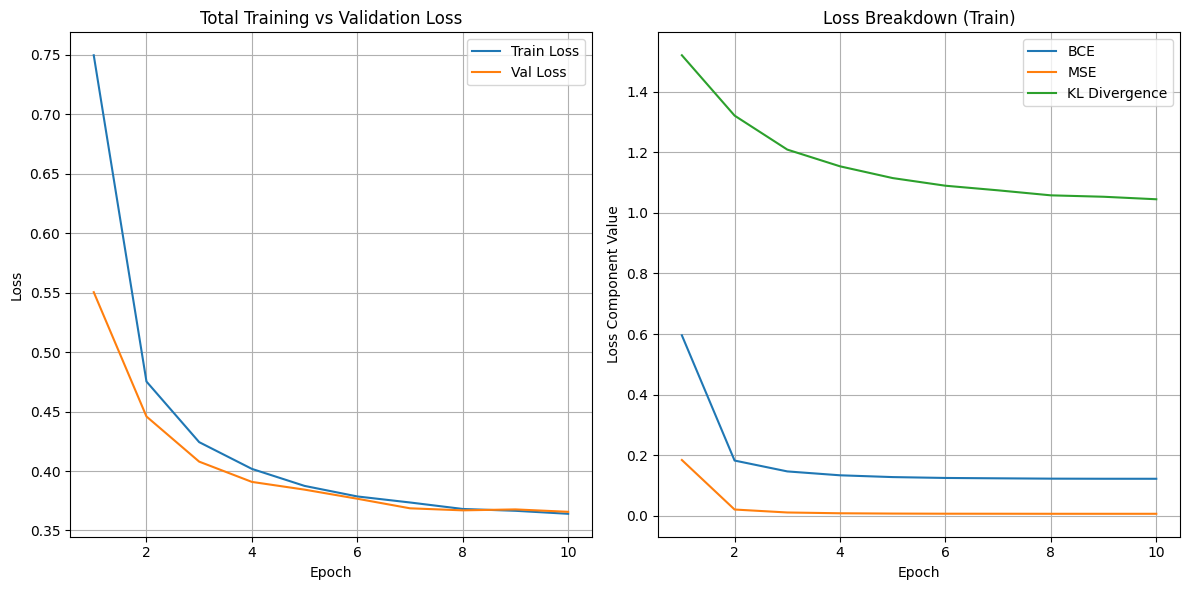}
    \caption{Training and validation loss curves across epochs. The composite loss includes BCE, MSE, and KL components. Close alignment between training and validation losses indicates strong generalization.}
    \label{fig:loss_curves}
\end{figure}

Quantitative training metrics confirmed convergence and generalization. Figure~\ref{fig:loss_curves} shows the evolution of the composite loss and its components-binary cross-entropy (BCE), mean squared error (MSE), and Kullback–Leibler (KL) divergence—over training epochs. Validation loss remained closely aligned with training loss throughout, suggesting effective regularization and minimal overfitting. 

While several standard metrics such as SSIM and Pearson correlation can be computed, they are not emphasized here due to the inherently stochastic and subject-dependent nature of human gaze. Pixelwise alignment may not reflect perceptual similarity or semantic agreement in fixation patterns. Therefore, we focus on qualitative visual agreement and spatial correspondence between predicted and observed fixation distributions.

These results confirm that the model effectively learns the spatial structure of fixation behavior without requiring explicit temporal information. The predicted heatmaps capture key aspects of human attention, including multimodality, salience-driven clustering, and variability across viewing conditions.

\subsection{Toward Practical Applications and Extensions}

The success of our heatmap-based framework highlights its strength as a generalizable and interpretable model of human gaze behavior. By treating visual attention as a spatial probability distribution, the approach accommodates non-stationary and subject-specific variability without requiring explicit temporal supervision. This enables training across diverse participants and viewing patterns while still preserving core visual saliency structure. The model’s outputs—dense fixation likelihood maps—offer direct insight into which regions are likely to attract attention, making them suitable inputs for downstream cognitive or semantic models of vision.

While the current formulation omits explicit temporal sequencing, it establishes a robust platform for future extensions. Temporal modules—such as lightweight RNNs or attention-based reordering schemes—can be layered on top to generate full scanpaths from predicted fixation regions. The model also lends itself naturally to real-world tasks like interactive attention tracking, adaptive AR interfaces, or early screening tools in clinical settings. In this way, our work not only bridges statistical and biological perspectives on eye movement, but also sets the stage for more temporally and behaviorally grounded models of human visual cognition.

\section{Discussion}

Human gaze behavior offers a unique window into the brain’s visual processing mechanisms. Operating within a $\sim$15W power constraint, the visual system demonstrates remarkable efficiency and adaptability. The statistical and modeling approaches in this study aim to reverse-engineer aspects of that efficiency through data-driven analysis.

Our results emphasize that human gaze is neither deterministic nor stationary. Instead, eye movements are stochastic, varying significantly across individuals. This raises key questions: Do people perceive the same image similarly? Is there a shared visual "language" in the brain, or is attention shaped primarily by experience and context? Defining geometric or probabilistic distances between fixation maps or scanpaths could quantify inter-subject variability and enable comparisons across individuals or populations.

Such metrics may uncover cognitive or perceptual trends, or highlight common semantic features that attract attention. Our finding that gaze patterns resemble Lévy walks—random walks with heavy-tailed step lengths—supports the hypothesis that visual attention is governed by optimal foraging principles. Dense clusters of fixations interleaved with long-range jumps suggest an efficient strategy for information sampling.

This inherent variability also limits the utility of traditional pixelwise metrics such as SSIM or Pearson correlation for evaluating fixation predictions\cite{bylinskii2018different}. Given that no two humans scan the same image identically, such measures may understate the true perceptual or semantic similarity between predicted and observed maps \cite{kummerer2015deep}.Consequently, our evaluation emphasizes qualitative alignment and spatial saliency structure.

This has clinical relevance: deviations from typical spatial or statistical gaze patterns could serve as early markers of neurological or perceptual disorders. Comparing predicted and observed fixation distributions across populations may assist in diagnosing conditions like autism, ADHD, or degenerative diseases.

In conclusion, modeling gaze as a spatial distribution prediction task provides a robust and interpretable alternative to autoregressive sequence modeling. Grounded in both statistical and biological insights, this framework opens avenues for future research in personalized gaze modeling, temporal extensions, and real-time human–AI interaction.

\section*{Acknowledgements}

I would like to express my sincere gratitude to Prof.\ Raghunathan Rengaswamy for his valuable advice and guidance throughout this work.  
I would also like to thank Lokesh Rajulapati for his fruitful discussions and helpful suggestions.  
Finally, I extend my gratitude to Prof.\ Rajagopalan Srinivasan, IIT Madras, Prof.\ Babji Srinivasan, IIT Madras, and the Data Analytics, Risk and Technology (DART) Lab for providing the facilities used to perform the experiments.




\appendix
\section{Appendix}
\includegraphics[scale=0.75]{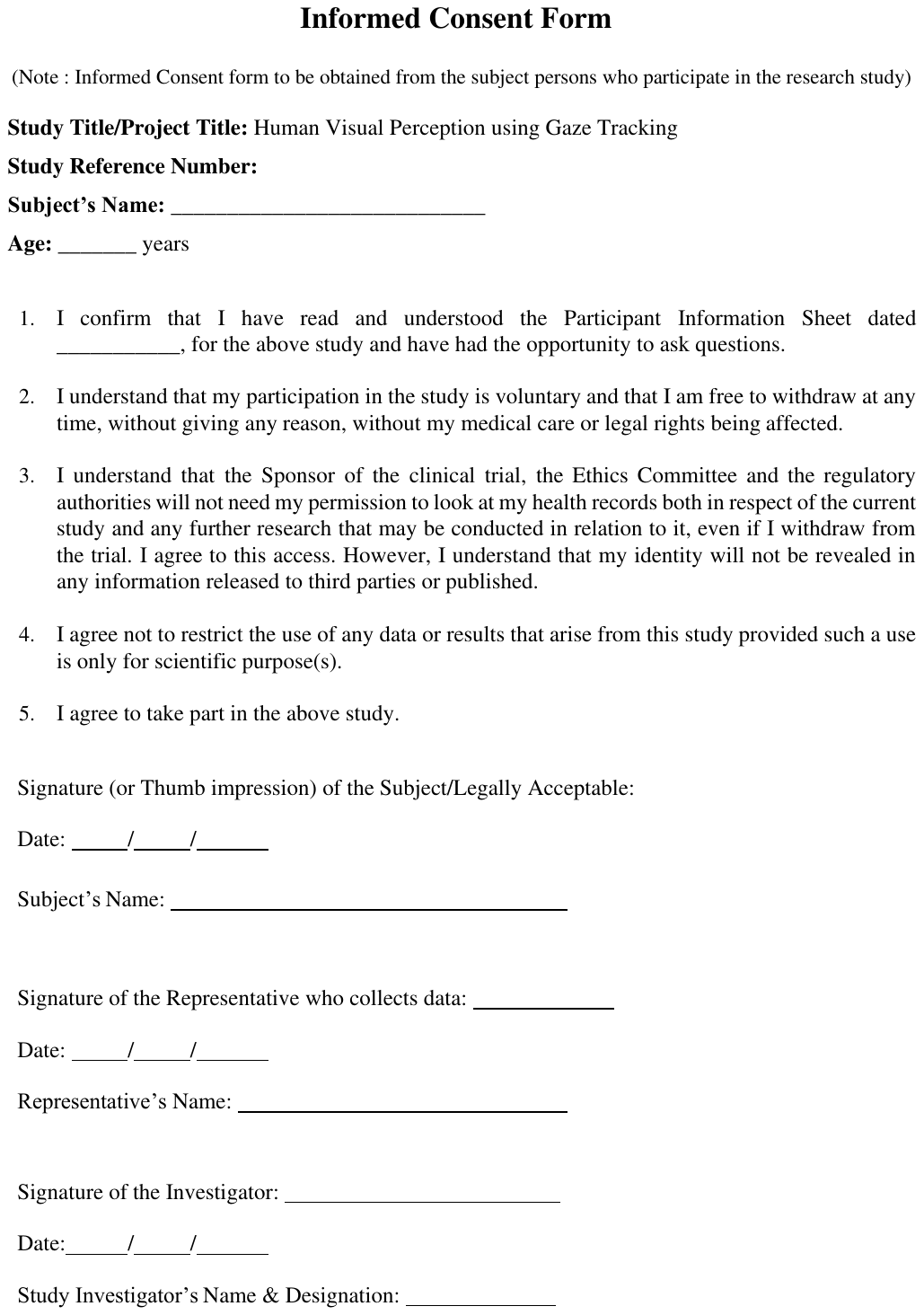}

\bibliography{references}
\bibliographystyle{plain}
\end{document}